\documentclass[conference]{IEEEtran}
\IEEEoverridecommandlockouts
\usepackage{cite}
\usepackage{amsmath,amssymb,amsfonts}
\usepackage{algorithmic}
\usepackage[dvipdfmx]{graphicx}
\usepackage{textcomp}
\usepackage{xcolor}
\usepackage{booktabs}
\usepackage{tabularx}

\usepackage{makecell}

\usepackage{multirow}
\usepackage{array}
\usepackage{url}
\usepackage{tikz}

\newcommand\copyrighttext{%
  \footnotesize \textcopyright 2026 IEEE. Personal use of this material is permitted. Permission from IEEE must be obtained for all other uses, in any current or future media, including reprinting/republishing this material for advertising or promotional purposes, creating new collective works, for resale or redistribution to servers or lists, or reuse of any copyrighted component of this work in other works.}
\newcommand\copyrightnotice{%
\begin{tikzpicture}[remember picture,overlay]
\node[anchor=south,yshift=10pt] at (current page.south) {\fbox{\parbox{\dimexpr\textwidth-\fboxsep-\fboxrule\relax}{\copyrighttext}}};
\end{tikzpicture}%
}

\def\BibTeX{{\rm B\kern-.05em{\sc i\kern-.025em b}\kern-.08em
    T\kern-.1667em\lower.7ex\hbox{E}\kern-.125emX}}
\begin{document}

\title{Script-to-Slide Grounding: Grounding Script Sentences to Slide Objects for Automatic Instructional Video Generation}

\author{

\IEEEauthorblockN{Rena Suzuki}
\IEEEauthorblockA{
\textit{Nagoya Institute of Technology}\\
Nagoya, Aichi, Japan \\
suzure@ozlab.org}
\and

\IEEEauthorblockN{Masato Kikuchi}
\IEEEauthorblockA{
\textit{Nagoya Institute of Technology}\\
Nagoya, Aichi, Japan \\
kikuchi@nitech.ac.jp}
\and

\IEEEauthorblockN{Tadachika Ozono}
\IEEEauthorblockA{
\textit{Nagoya Institute of Technology}\\
Nagoya, Aichi, Japan \\
ozono@nitech.ac.jp}
}

\maketitle
\copyrightnotice{}

\begin{abstract}
While slide-based videos augmented with visual effects are widely utilized in education and research presentations, the video editing process---particularly applying visual effects to ground spoken content to slide objects---remains highly labor-intensive.
This study aims to develop a system that automatically generates such instructional videos from slides and corresponding scripts.
As a foundational step, this paper proposes and formulates Script-to-Slide Grounding (S2SG), defined as the task of grounding script sentences to their corresponding slide objects.
Furthermore, as an initial step, we propose ``Text-S2SG,'' a method that utilizes a large language model (LLM) to perform this grounding task for text objects.
Our experiments demonstrate that the proposed method achieves high performance (F1-score: 0.924).
The contribution of this work is the formalization of a previously implicit slide-based video editing process into a computable task, thereby paving the way for its automation.
\end{abstract}


\section{Introduction}
Slide-based instructional videos, widely utilized in education and research, can be enhanced with visual effects such as pointing to deepen viewer comprehension; however, the significant effort required for their creation remains a challenge. 
To effectively facilitate audience understanding, visual effects---such as pointing and highlighting---must be synchronized with the corresponding figures, tables, and text on the slides with the narration.
Nevertheless, the manual process of adding these visual effects is exceedingly time-consuming and labor-intensive.
Furthermore, the quality of the final video is highly dependent on the editor's skills and interpretation, posing issues in terms of both efficiency and reproducibility.
The objective of this research is to develop a system that automatically generates slide-based videos augmented with visual effects. 
The system takes slides and a corresponding script as input, automatically adding visual effects based on the narration.
In this context, a script is the text that is narrated for the slides.
While existing research has addressed the automatic generation of slides and scripts, few studies have focused on the automatic application of the visual effects themselves—that is, 
determining ``what'' slide content to emphasize and ``when'' to do so based on the narration.

In this study, we deconstruct this automation process into two sequential tasks.
The first is a ``grounding'' task, which identifies the slide objects referenced by the script. 
The second, termed the ``Attention Control Problem,'' determines the specific visual effects to apply based on the grounding results.
This paper focuses on the former task, grounding, as it serves as the foundation for the entire process.
The accuracy of this grounding is fundamental to creating effective visual effects, and since the latter task is entirely dependent on the outcome of the former, it must be addressed first.
This grounding task is challenging as it requires semantic interpretation; despite recent advancements in vision-language models (VLMs), the unique structure of slides prevents their straightforward application.
This paper analyzes and formulates this fundamental task as Script-to-Slide Grounding (S2SG).
Automatically applying visual effects based on a script necessitates an integrated understanding of textual and visual information, which is a highly complex problem.
Therefore, we propose a phased approach that begins with the relatively simple task of grounding between textual elements, using these results as a foothold to tackle the more complex problem involving visual information.
In this paper, we introduce a method called Text-S2SG, which is limited to text-only slides, and demonstrate through evaluation experiments that it can achieve high-performance grounding (F1-score: 0.924).

\section{Related Work}\label{sec:refer}
The manual creation of slide-based videos is not only exceedingly time-consuming and labor-intensive but also poses challenges in efficiency and reproducibility, as the quality is highly dependent on the editor's skills.
Automating the application of visual effects based on a script is a formidable task, requiring an integrated understanding of textual and visual information.
We, therefore, adopt a phased approach: we first address the simpler task of grounding between textual elements, intending to use these results as a foundation for tackling the more complex problem of visual information.
To situate this approach, this section reviews existing research from three perspectives---support for creating slide-based videos, slide content analysis, and the grounding problem---to clarify the positioning of our study.

\subsection{Support for Creating Slide-based Videos}
Existing approaches aiming to automate the creation of slide-based videos can be broadly categorized into ``automated content generation'' and ``interactive editing support.''

In research on automated content generation, the focus has been on the automatic creation of slides and scripts. 
Xu et al. \cite{Taoxu_AIg}, in the context of English vocabulary learning, generated scripts from existing slide materials using a specialized large language model (LLM) and created lecture videos by vocalizing these scripts as narration. 
However, their work is limited to narration generation and does not extend to the application of visual effects. 
Ge et al. \cite{ge_autopresent} proposed a method for automatically generating slides solely from natural language instructions. 
Specifically, they designed a custom library that encapsulates basic slide creation operations such as image retrieval and title insertion. 
Using this library, an LLM generates a program to manipulate the {\it python-pptx} library, thereby creating slides with layouts and images.
However, this research is also focused on generating static slides and does not address the generation or application of visual effects.

In research on interactive editing support, LLM-based assistance has been proposed to reduce the workload of video editing. 
Wang et al. \cite{wang_lave2024} introduced ``LAVE,'' an agent system that understands and processes editing instructions in natural language to automatically plan and execute tasks such as clip selection, trimming, and sequence editing. 
However, the application of advanced visual effects is identified as a future challenge. 
``ExpressEdit'' by Tilekbay et al. \cite{tilekbay_expressedit2024} allows for the application of effects like shape overlays and zooming through a combination of natural language and sketches, but this is an interactive support system that relies on user instructions and does not achieve full automation of the editing process.

Creating slide-based videos augmented with visual effects to promote comprehension requires more than just the automation of material generation or editing operations.
It demands an advanced integrated process that dynamically corresponds narration with visual elements on the slides and designs and applies visual effects in a way that aids audience understanding. 
Gritz et al. \cite{Gritz_AIED25} reported that an increase in amount of text within a slide leads to an increase in comprehension-related actions such as pausing. 
This suggests that visually indicating the scope of slide objects corresponding to the narration can significantly contribute to comprehension. 
However, at present, no established approach exists to fully automate the editing process of determining ``what visual effect'' to apply, ``where,'' and ``when.''
Consequently, the creation of such videos remains dependent on manual labor, which not only entails a high workload but also suffers from reproducibility issues, as the final quality and style are dependent on the creator's subjectivity and skills. 
Therefore, establishing a mechanism that autonomously generates consistent visual effects from slides and a script is a critical challenge for the future of assistive technologies for creating slide-based videos.

\begin{figure}[tb]
    \centering
    \includegraphics[width=0.87\linewidth]{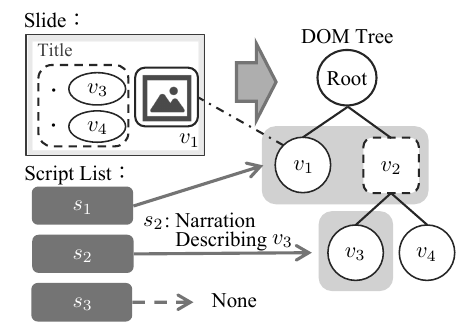}
    \caption{Overview of Script‑to‑Slide Grounding (S2SG)}
    \label{fig:problem}
\end{figure}

\subsection{Slide Content Analysis}
Accurately extracting meaning from slides is extremely difficult due to their unique data structure.
A presentation file (e.g., .pptx) is essentially an archive containing multiple XML files and other data.
The content of each slide is represented by a tree structure called a Document Object Model (DOM) tree, as shown in Fig. \ref{fig:problem}.
In this structure, individual slide objects such as text and shapes become nodes, and the entire slide is serialized as an XML document.

This DOM tree presents two problems for slide analysis: a lack of semantic structure and issues with object granularity. 

The first problem, the lack of semantic structure, arises because the DOM tree merely reflects the slide's creation history or internal data structure, not the semantic structure that humans perceive. 
For example, as shown in Fig. \ref{fig:problem}, a human might interpret the layout as indicating that node $v_2$ should be read before node $v_1$, but on the DOM tree, they could be recorded in the order $v_1$, $v_2$ depending on the editing sequence. 
As such, the order of appearance in the XML does not necessarily match the visual layout, so a naive approach that simply maps script sentences to objects sequentially cannot correctly reflect the presenter's intended structure. 
The second problem is object granularity. 
A semantic unit may be represented as a single object, rendering the granularity too coarse.
For instance, a bulleted list with indented items is often represented as a single text object. 
This poses a problem when processing the correspondence with a script, as it is difficult to accurately point to just one item in the list even if the script refers to it specifically.
Regarding the first challenge, Hayama et al. \cite{hayama_Ste} proposed a method that uses the visual layout information of a slide to identify the functional attributes of objects and extract a hierarchical structure. 
However, they report that analysis becomes difficult when the relationships between objects are defined by their text content. 
Furthermore, accurately understanding the meaning of non-text objects such as figures and tables requires considering the context and the entire slide's content.
This second challenge can be addressed by hierarchizing the text objects within a slide using indentation information, as seen in the research by Inoue et al. \cite{inoue_LS} and Sano et al. \cite{sano_SAM}. 
This approach aims to reflect the creator's intended layout and semantic divisions by analyzing the hierarchical structure of each text object. Meanwhile, VLMs are a technology applicable to slide content interpretation. 
VLMs are foundation models capable of simultaneously understanding and associating images and text. 
By inputting a slide as an image and a script as text, there is a potential to directly analyze the correspondence between them. 

However, significant challenges remain with this approach. 
Lee et al. \cite{lee_LPM2023} have shown that it is difficult for AI models to accurately understand the complex diagrams and, particularly, the mathematical formulas found in lecture slides, for which recognition accuracy is known to be low. 
As a result, they report that the grounding of speech to slide objects in lecture videos has not yet reached human-level performance. 
In other words, understanding the overall meaning of a slide remains a difficult task for VLMs.

\subsection{Slide Analysis as a Grounding Problem}

The challenge addressed in this research can be framed as a type of grounding problem, which involves connecting linguistic information to specific objects. We define the task of identifying which set of objects on a slide each sentence of a script refers to as S2SG.

Grounding is an active area of research, with studies such as ``Textual Grounding'' \cite{Cheng_ECCV24}, which extracts text regions within a document based on a query, and ``Visual Grounding'' \cite{LLaVA_ECCV24}, which associates specific regions in an image with text. 
However, directly applying these existing methods to S2SG is extremely difficult due to two unique characteristics of the slide medium.

The first is the semantic and pragmatic gap. Scripts frequently use abstract and deictic expressions (pragmatics), such as ``As this graph shows...'' In contrast, the corresponding object on the slide is a more concrete entity (semantics), like the entire graph or a trend within it. Bridging this gap in abstraction requires advanced reasoning capabilities that go beyond simple text matching or object detection.

The second is context dependency in a structured document. 
A slide is a structured document where objects like titles, bullet points, figures, and tables are intentionally arranged. 
The meaning of an object is not determined by the object in isolation but is strongly dependent on the overall layout structure of the slide and the flow of the presentation (context). Therefore, existing grounding methods that treat each object as independent cannot capture this context.

From the analysis above, S2SG can be considered a novel problem that cannot be solved by a mere extension of existing research. To address this complex problem, this paper proposes a phased approach that first uses an LLM, which excels at contextual understanding, to perform the relatively easier task of grounding between textual elements (script and on-slide text). 
This strategy aims to effectively simplify the overall S2SG problem by significantly reducing the search space for subsequent VLMs processing, thereby lowering the computational load and the risk of misinterpretation.

\section{Script-to-Slide Grounding}
This section describes S2SG, a task that involves estimating the correspondence between a slide and a script. 
First, we outline the problem setting and its associated challenges.
Then, we present the formal definition of S2SG and the method for representing the correspondence between slide objects and each sentence in the script (hereafter, a ``script sentence'').

\subsection{Problem Setting}\label{sec:setting}
We aim to build a system that automatically generates slide-based videos augmented with visual effects. 
Given a set of presentation slides and the corresponding script, an LLM estimates which slide objects should be visually emphasized based on the content of script sentences. 
The system must then estimate and associate the relevant slide objects for each script sentence and output the visual effect information to be applied during that sentence's playback.
This task can be divided into two sub-tasks: S2SG, which involves associating the script with slide objects, and the ``Attention Control Problem,'' which involves determining the visual effects for guiding the viewer's gaze. 
In this paper, we focus on S2SG.

In addition to the issues discussed in Section \ref{sec:refer}, several challenges arise.
First, there is a mismatch in information content and representation: slides are often more concise than scripts, and scripts may include abstract or deictic expressions (e.g., ``this'') and non‑verbal cues (e.g., pointing).

Second, the interpretation of graphical objects is context‑dependent; for example, even if a yellow triangle represents a banana, it is difficult to infer this unless the script explicitly states it. 
Accurately understanding non‑textual objects typically requires reasoning over the context and the slide’s overall structure. 
Finally, despite recent advances in image generation, precise highlighting over text‑heavy slides remains difficult to control; prompt design is costly, and processing time grows with the number of slides.

For these reasons, S2SG requires accurately estimating correspondence by leveraging semantic and structural relationships between script sentences and slide objects.

\subsection{Formal Definition}\label{sec:prb_teigi}
The objective of the S2SG task is to automatically estimate the correspondence between the script sentences in a script and the objects on a slide. 
In this study, we assume that the correspondence in S2SG is confined within a single slide.

For each slide, given a set of script sentences 
$S = \left\{s_i \right\}_{i=1}^{n}$,
and a set of slide objects 
$V =  \left\{v_j \right\}_{j=1}^{m}$,
we estimate a subset of slide objects $V_i = \left\{v \mid s_i \rightarrow v, \  v \in V \right\}$   
that corresponds to each script sentence $s_i \in S$. 
By these premises, 
the grounding function $g$ is defined as the following function:

\begin{equation}
g: s_i \rightarrow 2^V \text{.}
\label{eq:g}
\end{equation}

Here, $2^V$ represents the power set of the slide object set $V$. 
Let $g(s_i)$ be the set of slide objects estimated to correspond to the script sentence $s_i$; 
it can be empty if $s_i$ is unrelated to any slide objects (e.g., a digression).

Alternatively, the correspondence can be represented as a real-valued matrix $M$. 
The $M$ is formulated as:

\begin{equation}
M \in [0,1]^{n \times m} \text{.}
\label{eq:M}
\end{equation}

Here, $m_{ij}\in M$ indicates the likelihood that script sentence $s_i$ corresponds to slide object $v_j$,
with values near 1 indicating stronger correspondence.
Therefore, S2SG can be formulated as the problem of estimating this correspondence matrix $M$.

\section{Proposed Method: Text-S2SG}
This section proposes a Text-S2SG algorithm tailored for slides containing only text objects and 
describes a prototype system that incorporates this algorithm.

\subsection{Text-S2SG Algorithm}\label{subsec:syuhou}

\begin{figure}[tb]
    \centering
    \includegraphics[width=0.87\linewidth]{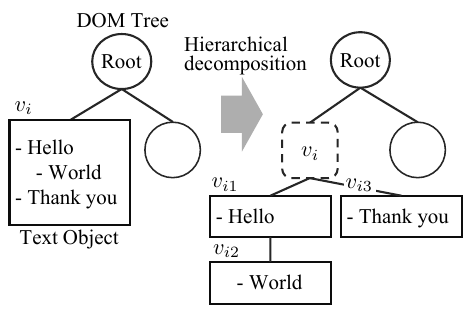}
    \caption{Hierarchical decomposition of text objects}
    \label{fig:method_textEle}
\end{figure}

In this paper, we propose an LLM-based method to automate S2SG under a simplified setting, where slides contain only text objects, in order to address the challenges mentioned in Section \ref{sec:setting} and to ensure experimental feasibility and clarity of evaluation.
Each script sentence---delimited by a period (.) or a newline---is treated as a unit. 
Slide text objects are split by newlines into sentence level elements.
The method consists of two stages: preprocessing the slide information and performing the grounding using an LLM.

As illustrated in Fig. \ref{fig:method_textEle}, each text object $v_i$ extracted from the slide's XML is split by newlines into $\{v_{i1}, \dots, v_{in}\}$. 
We then build a hierarchy by inferring parent-child relations from indent levels, thereby preserving the logical structure of the original text object $v_i$.

\begin{figure}[tb]
	\centering
	\fbox{%
	\begin{minipage}{0.9\linewidth}
		\small
		You are a presentation support assistant. For each sentence in the following presentation script, please ground it to the object it describes from the list of slide objects. 
		Note that not all objects will necessarily be described. Your output must be in JSON format. 
		For each sentence, return a list of shape\_IDs for the corresponding objects.
	\end{minipage}
	}
	\caption{Instruction prompt used for Text‑S2SG}
	\label{fig:prompt_eval}
\end{figure}

The processed slide information and the script are provided as input to the LLM, along with a pre‑designed Text-S2SG instruction prompt (Fig. \ref{fig:prompt_eval}).
For each script sentence, the model outputs the corresponding \texttt{shape\_ID}(s), a unique ID for a slide object.
To improve grounding accuracy, rules regarding context, group structures, and correspondence to multiple objects are also provided to the LLM. 
We believe that providing information on a per-page basis allows for a complete and concise presentation of the slide content.

The grounding in this method can also be represented as a binary decision on the relationship between each script sentence and slide object.
In the method proposed, $M$ can be treated as a binary matrix, where the entries are determined based on the LLM's output.
For example, when \( g(s_1) = \{v_2, v_4\} \) and \( g(s_2) = \{v_3\} \), the correspondence matrix $M$ would be represented as follows:

\begin{equation}
M =
\begin{bmatrix}
0 & 1 & 0 & 1\\
0 & 0 & 1 & 0
\end{bmatrix}
\text{.}
\label{eq:Mexample}
\end{equation}

With this design, we achieve the automation of the S2SG task for slides containing only text objects, while considering the logical structure of the slide layout and the information gap with the script's content.

\subsection{Prototype System}\label{sec:zissou}
\begin{figure}[tb]
    \centering
    \includegraphics[width=0.87\linewidth]{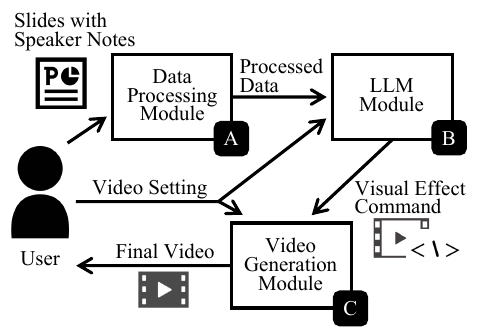}
    \caption{System architecture}
    \label{fig:arc}
\end{figure}

This prototype system takes slides with speaker notes as input and automatically generates slide-based videos augmented with visual effects. 
The system is limited to slide pages containing only text objects, and the speaker notes for each slide are used as the script. 
As shown in Fig. \ref{fig:arc}, the system consists of three components: (A) a Data Processing Module, (B) an LLM Module, and (C) a Video Generation Module.

(A) The Data Processing Module converts the slide presentation into a format usable by subsequent modules and simultaneously renders each slide as an image for video generation. 
First, it uses {\it python-pptx} to extract strings from each text object and assigns a unique identifier, \texttt{shape\_ID}, to each. 
If a text object contains multiple sentences, 
it is split by newlines, and each resulting line is treated as an independent sentence element, 
as described in Section~\ref{subsec:syuhou}. 
Then, a hierarchical structure is constructed by analyzing the indent level of each sentence element 
and determining parent-child relationships based on the indentation differences from the preceding sentences.
This hierarchy represents the logical parent-child relationships at the sentence level while preserving the structure of the original text objects (\texttt{ContentList}). 
Each text object has \texttt{\{group\_shape\_ID, ContentList\}} attributes, and each sentence element has \texttt{\{shape\_ID, Content, children\}} attributes.

Furthermore, it generates ``searchable slide data'' for the video rendering process. 
This data lists all slide objects on a page in a flat structure, with each object assigned \texttt{\{shape\_ID, Content, position\}} attributes. 
Here, \texttt{position} indicates the coordinate range on the slide image. The coordinates are assigned as follows: when converting the slide to an image via PDF, {\it PyMuPDF} is used to extract text and coordinate information. 
These results are then matched with the text information extracted by {\it python-pptx}.

\begin{figure}[tb]
	\centering
	\fbox{%
	\begin{minipage}{0.9\linewidth}
		\small
		You are a presentation video generation assistant. For each sentence in the following presentation script, please ground it to the object it describes from the list of slide objects. Note that not all objects are necessarily described. 
		The overall goal is to create a short video clip for each script sentence and then concatenate them to produce the final video. 
		Based on the grounding results, assign appropriate visual effect commands for each script sentence. You may assign multiple commands to a single sentence.
	\end{minipage}
	}
	\caption{Instruction prompt used for the LLM Module (B)}
	\label{fig:prompt_conduct}
\end{figure}

(B) The LLM Module takes the preprocessed slide data, the script, and a specific instruction prompt for this module (Fig. \ref{fig:prompt_conduct}) as input. 
It then outputs a list of corresponding \texttt{shape\_ID}s and visual effect commands for each script sentence. 
This output includes a simple implementation of the ``Attention Control Problem,'' allowing for the automatic application of visual effects based on the narration's content. 
The LLM is called on a per-slide basis (e.g., for a 3-slide presentation, slides 1, 2, and 3 are processed independently).

Each command output by the LLM has \texttt{\{type, start\_time, duration, param\}} attributes. 
The candidates for \texttt{type} are pointer, frame, and avatar,
which are represented as \texttt{POINT}, \texttt{RECTANGLE}, and \texttt{AVATAR}, respectively, in the command.
The pointer helps in guiding the viewer's gaze by dynamically pointing to the object being discussed, while the frame highlights a specific area to clearly indicate the object of attention. 
By providing these command candidates to the LLM, the system can automatically decide which object to emphasize, how, and when, according to the content of the narration. 
The meaning of each field is as follows. 
\texttt{start\_time} and \texttt{duration} represent temporal control within the short video clip, and \texttt{param} stores effect-specific parameters.
For example, for a pointer, \texttt{param = \{start\_pos, end\_pos\}}; for a frame, \texttt{param = position}.
The parameters \texttt{start\_pos}, \texttt{end\_pos}, and \texttt{position} store the \texttt{shape\_ID} 
of the corresponding object.

(C) The Video Generation Module generates a series of short video clips with visual effects rendered based on the LLM output. 
These clips are then sequentially combined to create the final video. 
Specifically, it uses the \texttt{shape\_ID} contained in each command to reference the coordinate information from the ``searchable slide data'' generated in step (A). 
For a frame command, for example, it draws a rectangle enclosing the coordinate range from  $(x_1, y_1)$ to $(x_2, y_2)$ for \texttt{duration} seconds, starting at \texttt{start\_time}. 
Furthermore, the user can configure the display position and visibility of the slide and avatar. 
When the avatar is set to be hidden, avatar-related commands are excluded from the prompt to ensure they do not appear in the LLM's output. 

\section{Evaluation Experiments}\label{sec:eval}
We evaluate the usefulness of Text-S2SG by examining how the amount and type of slide information provided to the LLM affects grounding accuracy.
From text-only slides, one can extract not only visible text but also coordinates, font sizes, and other attributes.
However, excessive input detail might hinder performance. We test this hypothesis using the Gemini 2.5 Flash model (temperature = 0).

\subsection{Experimental Method}
In this experiment, we collected academic presentation slides from four individuals and constructed a test set of 19 text-only pages (from 6 presentations) and 94 script sentences.
The text-only pages contained at least one textual element and may include decorative lines or arrows indicating the reading order, as well as headers, footers, or slide numbers, but excluded figures, tables, images, or other non-textual content.
The slides averaged $8.9$ objects (\textit{SD} $3.5$), scripts $4.9$ sentences (\textit{SD} $2.1$), uncorrelated; sentences averaged $46.4$ words (\textit{SD} $21.6$).
First, we considered two types of slide information: hierarchical (indent-based parent/child relations) and stylistic (font size, coordinates, role such as title/body).
We compared four data formats based on the presence/absence of these two types.
Next, we describe the evaluation methods. 
For the evaluation, we provided one page of slide information and a list of script sentences as a single input to the LLM, which then output a list of \texttt{shape\_ID} corresponding to each script sentence.
The instruction prompt used is shown in Fig. \ref{fig:prompt_eval}, along with the grounding rules mentioned in Section \ref{subsec:syuhou}.
After all outputs for every slide were generated, we compared the estimated results with the ground truth data and calculated the number of correct items for each script sentence.
A correct item was defined as a matching list element between the ground truth list and the estimated list.
For example, if the ground truth was [1,2,3] and the estimate was [1,2,4], the number of correct items would be 2. 
Based on this number, we calculated the F1-score for the entire set of test cases for each data format, using a micro-averaged approach (i.e., computed over the entire dataset).
The ground truth data was created subjectively by one of the authors.

\subsection{Experimental Results}
Table \ref{tab:result1} shows the experimental results.
Across four data formats (presence/absence of hierarchical and stylistic metadata), we observed no substantial differences;
the average F1-score was 0.924. 
Most mismatches involved erroneously selecting titles or parent/child items of a correct object; excluding titles from the targets can mitigate these cases.
Even when visual emphasis is applied based on such predictions, the discrepancies are unlikely to hinder audience comprehension.
This also suggests that, at least for text objects, the grounding process can be performed sufficiently well as long as the specific displayed content is provided.

\begin{table}[t]
\caption{F1‑scores for each data format} 
\label{tab:result1}
\centering
\renewcommand{\arraystretch}{1.15}
\begin{tabular}{@{}>{\centering\arraybackslash}p{2.5cm}
                  >{\raggedleft\arraybackslash}p{1.5cm}
                  >{\raggedleft\arraybackslash}p{1.5cm}
				  >{\raggedleft\arraybackslash}p{1.5cm}@{}}
\toprule
\multirow{2}{*}{\makecell[c]{\textbf{Stylistic info}}} 
& \multicolumn{2}{c}{\textbf{Hierarchical info}} \\
\cmidrule(l){2-3}
 & \multicolumn{1}{c}{\textbf{Present}} & \multicolumn{1}{c}{\textbf{Absent}} & \multicolumn{1}{c}{\textbf{Average}} \\
\midrule
\textbf{Present} & 0.923 & 0.930 & 0.926 \\
\textbf{Absent}  & 0.928 & 0.916 & 0.922 \\
\midrule
\textbf{Average}  & 0.925 & 0.923 & 0.924 \\
\bottomrule
\end{tabular}
\end{table}

\section{Discussion}

The S2SG proposed in this study is significant as it is the first to explicitly define the video editing work, which has hitherto been performed implicitly, as a scientifically verifiable task.
The high performance (F1-score: 0.924) on text-only S2SG achieved in our evaluation experiments demonstrates that the novel challenge of S2SG can be solved by an LLM.
This is a crucial achievement that paves the way toward the ultimate goal of automatically applying visual effects.
A particularly interesting finding from the experimental results is that the presence or absence of ``stylistic information,'' such as coordinates and font sizes, did not affect accuracy. 
Randomly shuffling the script sentences had no effect on performance.
This strongly suggests that the LLM makes its judgments based on the content (semantics) of the text rather than its superficial appearance. 
This discovery offers a promising outlook for future system design, indicating that high performance can be expected from a simple, text-centric approach without necessarily handling complex visual features.

Furthermore, our method's content-based (rather than order-based) grounding suggests that it is, in principle, capable of handling non-linear discourse structures, such as when a presenter moves back and forth between topics. 
Additionally, the observation that mismatches occurred in semantically related areas like titles and parent-child objects indicates that the method already captures semantic proximity at a high level, implying that further accuracy improvements can be expected with the addition of simple rules. 
These points illustrate the high potential for the advancement of this approach.

This paper has only addressed text objects, which is merely the first step in the broader S2SG problem. 
However, this success is significant because it validates the ``phased approach'' we proposed: first, achieve high-performance grounding between texts, and then use these results as a foothold to tackle the more complex problem of grounding non-text objects (e.g., by leveraging VLMs). 
Presenting this feasible development roadmap is one of the major contributions of this research.

The high-performance grounding established in this study is merely the first step in video generation. 
The next challenge is to solve the more advanced ``Attention Control Problem'': determining which objects to emphasize, at what timing, and with which 
visual effects, based on these grounding results. 
Findings from cognitive science indicate that inappropriate visual effects can increase the cognitive load on learners and hinder their understanding \cite{Sweller_J}. 
Therefore, the selection of which groundings should receive effects and the scheduling of when to display them become new research challenges that must be solved integrally to maximize learner comprehension. 
The results of this study provide the indispensable foundational technology for tackling this complex problem, and we are confident that it will lead to a future where anyone can easily produce high-quality educational and research content.

\section{Conclusion}
In this study, we aimed to automate the creation of slide-based videos with visual effects and formulated S2SG as a foundational task. 
We developed Text-S2SG for text-only slides and demonstrated its effectiveness, achieving an average F1-score of 0.924 across multiple data formats without notable performance differences. 
Qualitative observations further indicated that most errors involved semantically related objects, which are unlikely to hinder viewer comprehension.

These findings validate a phased approach that first focuses on text grounding.
Future work will extend the method to slides containing non-textual elements and address the ``Attention Control Problem'' to determine appropriate visual effects.

\section*{Acknowledgment}
This work was supported in part by JSPS KAKENHI Grant Numbers JP24K03052, JP25K21351.

\bibliographystyle{IEEEtran}
\bibliography{references3}

\end{document}